\documentclass[manuscript,screen]{acmart}
\usepackage{booktabs}
\AtBeginDocument{%
  \providecommand\BibTeX{{%
    \normalfont B\kern-0.5em{\scshape i\kern-0.25em b}\kern-0.8em\TeX}}}

\setcopyright{acmcopyright}
\copyrightyear{2018}
\acmYear{2018}
\acmDOI{XXXXXXX.XXXXXXX}

\acmJournal{JACM}
\acmVolume{37}
\acmNumber{4}
\acmArticle{111}
\acmMonth{8}

\acmPrice{15.00}
\acmISBN{978-1-4503-XXXX-X/18/06}

\begin{document}

\title{Analyzing the Effect of Sampling in GNNs on Individual Fairness}

\author{Rebecca Salganik}
\affiliation{%
  \institution{Universite de Montreal/MILA}
  \city{Montreal}
  \state{Quebec}
  \country{Canada}
}
\email{rebecca.salganik@umontreal.edu}

\author{Fernando Diaz}
\affiliation{%
  \institution{Google/MILA}
  \city{Montreal}
  \state{Quebec}
  \country{Canada}
}
\email{diazf@acm.org}

\authornotemark[1]
\author{Golnoosh Farnadi}
\authornotemark[1]
\affiliation{%
  \institution{HEC/Universite de Montreal/MILA}
  \city{Montreal}
  \state{Quebec}
  \country{Canada}
}
\email{farnadiq@mila.quebec}

\renewcommand{\shortauthors}{Salganik, Diaz, and Farnadi}

\begin{abstract}
  
In recent years, graph neural network (GNN) based methods have saturated the field of recommender systems. The gains of these systems have been significant, showcasing the advantages of interpreting data through a network structure. However, despite the noticeable benefits of using graph structures in recommendation tasks, this representational form has also bred new challenges which exacerbate the complexity of mitigating algorithmic bias. When GNNs are integrated into downstream tasks, such as recommendation, bias mitigation can become even more difficult. Furthermore, the intractability of applying existing methods of fairness promotion to large, real world datasets places even more serious constraints on mitigation attempts. As such, our work sets out to fill in this gap by taking an existing method for promoting individual fairness on graphs and extending it to support mini-batch, or sub-sample based, training of a GNN, thus laying the groundwork for applying this method to a downstream recommendation task. We evaluate two popular GNN methods: Graph Convolutional Network (GCN), which trains on the entire graph, and GraphSAGE, which uses probabilistic random walks to create subgraphs for mini-batch training, and assess the effects of sub-sampling on individual fairness. We implement an individual fairness notion called \textit{REDRESS}, proposed by Dong et al., which uses rank optimization to learn individual fair node, or item, embeddings. We empirically show on two real world datasets that GraphSAGE is able to achieve, not just, comparable accuracy, but also, improved fairness as compared with the GCN model. These finding have consequential ramifications to the fields of individual fairness promotion, GNNs, and in downstream form, recommender systems, showing that mini-batch training facilitate individual fairness promotion by allowing for local nuance to guide the process of fairness promotion in representation learning. 
\end{abstract}

\ccsdesc[500]{Computing methodologies~Structured outputs}
\ccsdesc[300]{Computing methodologies~Batch learning}
\keywords{Graph Neural Networks, Algorithmic Fairness, Individual Fairness}

\maketitle

\section{Introduction}
As our consumption habits shift towards online spaces, recommender systems are slowly becoming the gatekeepers between producers and consumers within a wide array of domains. In recent years, the field of recommender systems has become saturated with one particular form of representation learning: graph neural networks (GNNs) \cite{GNN_in_RecSys, GNN+RS_SURVEY2, GNN+RS_SURVEY3, GNN+RS_SURVEY4, Dau2019_GNN+RS_SURVEY5, Chicaiza2021ACS_GNN+RS_SURVEY6}. The key advantage which distinguishes GNNs from other representation learning methods is their ability to leverage not just the information embedded in the features associated with each data point, but also the information which can be extracted from their interactions \cite{GNN_survey}. However, in parallel to the proliferation of this important architecture, has come a rising level of concern that GNNs come with their own set of unique challenges when it comes to mitigating algorithmic bias. Findings indicating perpetuation or exponentiation of societal biases at the hands of other deep learning models have prompted similar questions in the field of GNNs, specifically as applied to recommender systems \cite{GNN_bias}. As such, the integration of these architectures into recommender systems can have very real, negative consequences on the experiences of both producers and consumers who interact with the recommender system. Without developing bias mitigation strategies for GNN's we cannot properly ameliorate the biases in downstream tasks, such as recommendation. 

Within the algorithmic fairness community, there have been several major thrusts, each associated with a broad definition of what it means to engage with fairness from a mathematical perspective \cite{Dwork_indfairness, Kusner_counterfactF, Kleinberg_groupF1}. The focus of this work is centered on the notion of individual fairness \cite{Dwork_indfairness}. This notion dictates that ''similar individuals should be treated similarly by an algorithmic system``. Intuitively, promotion of this notion can begin to address problematic discrepancies between the treatment of different consumer and producer groups in the recommendation space \cite{ekstrand}. Recent works that apply individual fairness notions to GNNs take different perspectives on similarity \cite{Lahoti_PFR, Kang_inform, Dong_ReDress}. In particular, \textit{REDRESS}, one such fairness framework, proposed by \citet{Dong_ReDress}, frames individual fairness from a ranking perspective. This makes the approach particularly appealing to integration with a downstream task of recommendation. 

\indent Our main contribution addresses one of the key limitations of REDRESS \cite{Dong_ReDress}. The use of a Graph Convolutional Network (GCN) \cite{Kipf16_GCN} as a backbone GNN model severely limits the ability of this method to scale to large datasets. As such, our work expands the proposed REDRESS framework to support sub-sampling methods and assesses the effects of this subsampling on the global individual fairness of the final embeddings. We re-implement the methodology proposed by Dong et al. and add our own implementation of a modified GraphSAGE \cite{Hamilton17_GS} which is compatible with individual fairness promotion. Our empirical evaluation on two real world datasets shows that the use of subsampling can drastically improve the individual fairness exhibited in the final embeddings. In doing so, we uncover a connection between neighborhood selection and individual fairness promotion.

\section{Related Work}
In this section we lay the groundwork for the various facets of work which play a crucial role in shaping the contribution and relevance of ours. First, we introduce the role that GNNs play in recommender systems. In doing so, we motivate the relevance of our approach to the recommendation space. Then, we introduce the need for feasible individual fairness notions which can be applied to ameliorate issues of bias in recommender systems. In this discussion, we highlight the ways in which our specific fairness framework, REDRESS \cite{Dong_ReDress}, can be applied to a recommendation setting. We follow this with a discussion of other individual fairness frameworks that have been applied to GNNs. Our discussion contrasts the other approaches with REDRESS, showcasing its intuitive connections with item-item relevance. Finally, we introduce the importance of sub-sampling in GNN mini-batch training. We position our work in relation to other surveys which explore the relation between sub-sampling and utility gains. In doing so, we highlight the crucial contribution of our work: the connection between sub-sampling and individual fairness promotion.

\noindent \textbf{GNNs in Recommender Systems}: One of the main factors which cements the success of GNN techniques in the recommendation domain is their ability to capture both context and content based patterns \cite{GNN_in_RecSys}. This enables GNNs to encode crucial collaborative signals embedded in a network of users and/or items as well as individual information about items and users themselves. The basis of taste and affinity for an item can be expressed both contextually, via our social circles, and individually, via our interactions with the item itself. As such, GNNs ability to capture both these sources of information enable them to build profoundly complex and robust representations of users and items. 

\indent The motivation behind our selection of GraphSAGE \cite{Hamilton17_GS} as a comparison method with the original GCN architecture proposed by \citeauthor{Dong_ReDress} is its achievements within the recommender system domain. PinSage \cite{pinSAGE}, proposed by the authors of GraphSAGE one year after its initial release, is designed as a downstream extension of GraphSAGE to recommendation tasks. In terms of its architecture, PinSage remains essentially the same as GraphSAGE, with a slight modification to the neighborhood selection method, which uses \textit{importance pooling} to refine the random walks based on the probability of a node's occurrence. After generating item and used embeddings using the iterative representation learning procedure of GraphSAGE, PinSage performs recommendation based on the k-nearest neighbors of an item or a user in the learned embedding space. As such, by implementing fairness in GraphSAGE, we lay the foundations for easily extending this approach to the recommendation task via PinSage. 

\noindent \textbf{The Need for Individual Fairness in Recommender Systems}:
One of the major issues which plagues the field of recommendation is popularity, or exposure bias \cite{GNN_bias}. This phenomenon occurs when users are only exposed to a portion of the available items. Frequently, this exposure is prioritized in the favor of previously popular items, leaving niche or new items to be neglected in negative feedback loop which catapults some items to \textit{superstar} popularity \cite{bauer_music_2017} and leaving others languishing in obscurity. This issue has a plethora of negative downstream consequences. First, it deteriorates the overall quality of predictions if a recommender system is able to serve only the needs of a mainstream audience. Second, it can create unfairness among consumers since not all consumption reflect mainstream tastes \cite{ekstrand}. Finally, it can have very real, extremely negative consequences on the financial prospects of producers \cite{popularity_Bias}, affecting their ability to create content, services, or merchandise. 

\indent Despite a wide range of works approaching exposure and popularity bias from the perspectives of group fairness \cite{Li_2021_group+RS, ekstrand} and counterfactual fairness \cite{Li_2021_CF+RS, Zhang_2021_C+RS2}, the granularity of individual fairness makes the promotion of this notion significantly more difficult. One of the benefits of the REDRESS \cite{Dong_ReDress}, the approach we incorporate into our work, is its flexibility in defining what features/structure is used to define similarity among items (see Section \ref{fair_training} for more details). For example, in the musical domain, if two songs sound extremely similar then, by selecting the portion of their feature set that defines their sound qualities, we can promote similarity between the two songs and regularize for inconsistencies based on the popularity of their respective primary artists. 



\noindent \textbf{Individual Fairness in Graphs}: In the broader discussion of fairness in machine learning, the tasks of promoting fairness is often broken into the three stages of a training pipeline: pre-processing (model), in-processing (during training, through regularization, constrained optimization, or novel loss functions), and post-processing (outcome generation). Fairness in graph representation learning follows a similar break down: fairness in the initial graph \cite{Kang_inform}, fairness in the mining (or representation building) algorithm \cite{bose19_advFinG, Lahoti_PFR, arduini_advfairness, Dong_ReDress}, and fairness in the final representations (or other downstream outcomes) \cite{Kang_inform}. Lastly, the approaches used can also be loosely categorized based on their fairness notions: group \cite{Chierichetti_fairlets, bose19_advFinG, ModelingBias_Dong_2022, arduini_advfairness}, individual \cite{Dong_ReDress, Lahoti_PFR, Kang_inform}, and counterfactual \cite{Ma_counterfacFinG} fairness techniques.


Since the focus of this work is individual fairness, we will now detail some of the most relevant individual fairness notions for GNNs. 
In their work, Kang et al. convert the premise of a Lipshitz condition into the graph domain \cite{Kang_inform}. They define individual fairness as a trace maximization problem bounded by a fairness tolerance, $\delta$. They break the task of promoting individual fairness into three steps offering three different approaches to satisfy the notion defined above: 1. Pre-processing the input graph 2. In-processing within the model, or 3. Applying post processing on the generated representations. However, due to its reliance on the graph laplacian for defining fairness, unlike our selected method, REDRESS \cite{Dong_ReDress}, the methodology proposed by Kang et al. does not lend itself well to the recommendation setting due to the nature of their fairness definition. Furthermore, in their paper, Dong et al. show that REDRESS outperforms InFoRm in empirical studies \cite{Dong_ReDress}. 

Meanwhile, \citeauthor{Lahoti_PFR} take a different understanding of individual fairness \cite{Lahoti_PFR}. Their Pairwise Fair Representation (PFR) model uses a sparse fairness graph with expert defined pairwise similarities to learn fair representations. Similar to the previous method, the methodology proposed by Lahoti et al does not easily lend itself to the recommendation setting. Unlike REDRESS which is able to perform both link prediction and node classification, Pairwise Fair Representations (PFR) is an unsupervised learning method which is unable to perform link prediction. 


\noindent \textbf{Sampling Methods in GNNs}: Despite the significant advances achieved by the GCN architectures, there are a few drawbacks to this method. The training regime used by this architecture, which performs full-batch training on the entire graph, has two major limitations: efficiency and scalability. Holding an entire graph in memory whilst performing complex iterative convolutions makes training a GCN on large graphs essentially intractable \cite{sampling_survey1}. As such, a series of modifications have been proposed which enable the use of mini-batch training while maintaining the expressive properties of a GCN's convolutional representation building \cite{Hamilton17_GS, pinSAGE, ASGCN, FastGCN, LADIES, ClusterGCN, GraphSAINT}. The variety of these sampling methods and their effects on accuracy and performance have been well documented both empirically \cite{sampling_survey1, sampling_survey2} and theoretically \cite{samplesurvey3}. As mentioned in our introduction of the REDRESS \cite{Dong_ReDress} framework, and explored with more detail in Section \ref{sec:methodology}, the primary drawback of this method is its reliance on the GCN methodology. By exploring the effects of subsampling with respect to individual fairness, we differentiate our findings from the various surveys that have assessed subsampling through the lens of utility performance. 


\section{Graph Neural Networks}
In this section, we review several key preliminary topics which are necessary for the discussion for understanding our discussion of REDRESS in Section~\ref{sec:methodology}. Advances in the field of GNNs have come in leaps and bounds within the last decade. Here, we briefly review the two methods integral to our work: GCN \cite{Kipf16_GCN} and GraphSAGE \cite{Hamilton17_GS}. For a detailed overview of GNN methods, we refer to the survey paper by Wu et al. \cite{GNN_survey} and a textbook by Hamilton \cite{Will_book}. 




\newcommand{\updatefn}[0]{\textit{UPDATE}} 
\newcommand{\aggfn}[0]{\textit{AGGREGATE}} 

Given graph data, a GNN iteratively collects feature information from the neighbors of a node and integrates this information into the representation of the node. As such, the basic structure of a GNN can be abstracted into two functions, $\updatefn$ and $\aggfn$. Here, $\aggfn$ refers to the process of gathering the neighborhood nodes and $\updatefn$ refers to the procedure used to update a node's embedding with respect to its neighborhood feature set. Thus, the GNN process of training embeddings, is based on the interactions between these two processes. At each iteration $k$ of training, the $\aggfn$ function takes the embeddings of $u$'s neighbors, $v \in N(u)$, to generate a message which the $\updatefn$ function combines with the previous embedding of node $u$, $h_u^{(k-1)}$, to generate the updated embedding $h_u^{(k)}$. This process can be formalized as follows:

\begin{align}
     h_u^{(k)} = \updatefn^{(k-1)}\biggl(h_u^{(k-1)}, \aggfn^{(k-1)}\left(\{h_v^{(k-1)}, \forall v \in N(u) \}\right)\biggr)
\end{align}
More formally, this can be expressed as: 
\begin{align}
    h_u^{(k)} = \sigma\biggl(W_{self}^{(k)}h_{u} ^ {(k-1)} + W_{neigh}^{(k)} \sum_{v \in N(u)}h_v^{(k-1)} + b^{(k)} \biggr)  
\end{align}
where $W_{\text{self}}^{(k)}, W_{\text{neigh}}^{(k)} \in \mathbb{R}^{d^{(k)} \times d^{(k-1)}}$ are trainable parameter matrices, $b^{(k)}$ is the bias term, and $\sigma$ is a non-linear activation function (in our experimentation we use ReLU \cite{relu}). 

\subsubsection{GCN}
The GCN method proposed by Kipf and Welling \cite{Kipf16_GCN} is unique because of its use of neighborhood normalization. Rather than simply aggregating over all the feature vectors of the neighborhood nodes, which can lead to instability, the neighboring feature set is first normalized. 

\begin{align}
    h_u^{(k)} = \sigma\biggl(W^{(k)} \sum_{\{v \in N(u)\} \bigcup \{u\}} \frac{h_v}{\sqrt{|N(u)||N(v)|}}\biggr)
\end{align}
\subsubsection{GraphSAGE} \label{GS}
GraphSAGE \cite{Hamilton17_GS} makes two major changes to the methodology proposed by GCN: neighborhood selection and aggregation. First, unlike GCN which integrates information from the all the neighbors of a node, GraphSAGE randomly samples \textit{p} neighbours from the surrounding neighborhood (i.e., $N_p(u)$). Second, unlike GCN, which normalizes based on the degree of the neighborhood nodes, GraphSAGE averages the embeddings of the neighborhood feature set before integrating them with the representation of the current node. As such, this can be expressed as: 
\begin{align}
    h_u^{(k)} = \sigma\biggl(W^{(k)} \cdot \text{MEAN}\Bigl(\{h_u^{(k-1)}\} \bigcup \{ h_v^{(k-1)}, \forall v \in N_p(u)\}\Bigr)\biggr)
\end{align}
where the $\text{MEAN}$ aggregator is the elementwise mean over the feature vectors, $h_v^{(k)}$ for all the neighboring nodes, $v \in N(u)$.


\section{Rank Based Individual Fairness in GNNs} \label{sec:methodology}

In this section, we review the individual fairness notion called \emph{REDRESS} which is introduced by Dong et al. \cite{Dong_ReDress}. We then show how REDRESS can be implemented within the end-to-end training paradigm of a GNN.

\subsection{REDRESS} \label{fairness_notion}

Our problem definition is couched in the graph setting. We are given a graph, $G = (V, E)$, with a node set, $V$, and edge set, $E$. Our setting also involves a set of node-level features, $X \in  \Re^{|V| \times d}$ where $d$ is the dimension of each feature vector. Our current task involves training a GNN model, $\mathcal{M}$, to perform the task of link prediction. In this setting, $Y$ and 
$\hat{Y}$ represent the ground truth and learned node representations. However, this methodology can easily be extended to the recommendation setting by treating the nodes $v \in V$ as items and attempting to predict items which are similar to each other, while maintaining individual fairness among them. By training a model, $\mathcal{M}$ on graph, $G$, we can learn a series of representations for each node, or item, $\hat{Y} \in \Re^{|V| \times m}$ where $m$ represents the dimension of the final hidden layer in model, $\mathcal{M}$. 

\indent The fairness notion used in this work is based on the framework proposed by Dong et al. \cite{Dong_ReDress}. This fairness notion, which bases itself on ranking, is extremely well suited to recommendation tasks due to its innate interlacing with item to item relevance. We begin by explaining this notion intuitively and follow with a formal definition: 

\indent For each node pair $u, v$ in a graph $G$, we can define their similarity, $s_{u,v}$ based on some similarity metric $s(., .)$ in which we input their initial feature vector, $X[v] \in \Re^{d}$ or learned embedding, $\hat{Y}[v] \in \Re^{m}$. Applying this procedure in a pairwise fashion, we can derive two matrices. The first is termed the \textit{apriori} or \text{oracle similarity}, $S_G$, in which similarity is based on some initial feature based, structural, or expert defined vectors which are independent of the GNN model, $M$. The second is termed the \text{learned representation similarity}, $S_{\hat{Y}}$, in which similarity is based on a learned embedding generated by the training process of a GNN model, $M$. Using these similarity matrices, we can generate two lists which rank each node with respect to the others. Individual fairness is upheld in this setting if the respecting orderings of nodes in each ranked list is consistent. That is, items which were ordered based on similarity before training, will maintain the same similarity ordering in their embedding space. Formally, we can express this in the following definition: 
\begin{definition}(REDRESS: Individual fairness). Given some similarity metric $s(., .)$, an apriori pairwise similarity matrix, $S_G \in \Re^{|V| \times |V|}$, and a learned pairwise similarity matrix, $S_{\hat{Y}} \in \Re^{|V| \times |V|}$, defined by applying $s(.,.)$ to the learned graph representations, $\hat{Y}$, we say the predictions are individual fair if for each each instance $i$, the ranked list generated from $S_{\hat{Y}}$ is consistent with the ranked list generated from $S_G$. 

\end{definition}
\subsection{REDRESS Fairness Construction} 

Minimizing the differences between the two ranked lists can be seen as a form of ranking optimization. In order to formulate this task as a differentiable operation, we can draw on the work in the field of learning to rank \cite{LTR} to formulate loss based on a probabilistic approach. Using these matrices, $S_G$ and $S_{\hat{Y}}$, we define two probability matrices, $P \in \Re^{|V| \times |V| \times |V|}$ and $\hat{P} \in \Re^{|V| \times |V| \times |V|}$. In essence, for every node $u_i$, we can define its similarity with two neighboring nodes $u_j, u_m$ as $s_{i,j}$ and $s_{i, m}$ in the case of $S_G$ or $\hat{s}_{i, j}$ and $\hat{s}_{i,m}$ in the case of $S_{\hat{Y}}$. Using these values, the matrices $P$ and $\hat{P}$ codify the probability $P_{j,m}(s_{i,j}, s_{i,m})$ that node $u_i$ is more similar to node $u_j$ than $u_m$.

Thus, for each individual node, $u_i$ and two other nodes selected from the remaining graph, $u_j, u_m$, the embedding based matrix, $\hat{P}$ is defined as: 
\begin{center} 
    $\hat{P}_{j,m}(\hat{s}_{i,j}, \hat{s}_{i,m}) = \frac{1}{1 + \exp\Bigl({-\alpha(\hat{s}_{i,j}-\hat{s}_{i,m)}}\Bigr)}$
\end{center}
where $\alpha$ is a scalar which can be treated as a hyperparameter.  
Similarly, the feature based matrix, $P$ can be defined as: 
\begin{center}
    $P_{j,m}(s_{i,j},s_{i,m})= \begin{cases}
  1  & s_{i,j} > s_{i,m} \\
  0.5 & s_{i,j} = s_{i,m} \\ 
  0 & s_{i,j} < s_{i,m}
\end{cases}$
\end{center}
\subsection{Training Individually Fair Embeddings} \label{fair_training}
In defining these matrices, we can now train for fairness by using standard cross entropy loss. For a node $i$, we define: 
\begin{align}
    L_{j,m}(i) &= -P_{j,m} \log \hat{P_{j,m}} - (1 - P_{j,m}) \log (1 -  \hat{P_{j,m}}) 
\end{align}
And over all the nodes $i \in V$: 
\begin{align}
    L_{fairness} = \sum_{i} \sum_{j,m} L_{j,m}(i)
\end{align}
Finally, in order to ease the computational complexity, we can add a weighing factor based on the changes in NDCG to restrict our modifications to changes in the top k values of each ranked list. Once we define $z_{@k}(., .)$ as NDCG@k, we can express the fairness loss as: 
\begin{align}
    L_{fairness} = \sum_{i} \sum_{j,m} L_{j,m}(i) |\triangle z_{@k}|_{j,m}
\end{align}
where $\triangle$ represents the change in NDCG between the two lists. 

\indent The broad process of training occurs using two stages: utility based loss and utility + fairness based loss. The utility loss, $L_{utility}$, is the classic cross entropy loss: 
\begin{align}
    L_{utility} = - \sum_{u,v \in G} Y_{uv}ln\hat{Y}_{uv}
\end{align}
and the joint utility + fairness loss can be defined as: 
\begin{align}
    L_{total} = L_{utility} + \gamma L_{fairness}
\end{align}
where $\gamma$ is a scalable hyperparameter which controls the focus given to fairness. By using the NDCG metric, commonly used in ranking and recommendation tasks, REDRESS formulates a fairness loss that is, intrinsically, performing the relevance/fairness balance that is so important to developing novel bias mitigation techniques in recommendation settings. Furthermore, by creating a loss which can easily interpolate between utility, or relevance, and fairness, REDRESS follows methodology very similar to a large body of work in recommendation fairness \cite{Wu2021MultiFRAM, marketplace}. 
. 

\section{Experiments}

In this section, we evaluate two implementations of REDRESS, comparing the results between GCN and GraphSAGE as the backbone model, $\mathcal{M}$. This section shows the effectiveness of sub-sampling on utility and fairness through empirical evaluations on two widely known real world benchmark datasets: BlogCatalog \cite{Blog_Catalog} and Flickr \cite{Flickr}. Table~\ref{datastats} presents the statistics of these two datasets. \\
The code and data used to obtain these results can be found at \href{https://github.com/Rsalganik1123/FaccTRec2022}{FacctRec2022 Github Repository}. 

\begin{table}[H] 
\centering
\footnotesize
\begin{tabular}{rccc}
\hline
\textbf{Dataset} & 
      \# Nodes & 
      \# Edges & 
      \# Features  \\
\hline 
BlogCatalog \cite{Blog_Catalog}&
      5,196 & 
      171,743 & 
       8,189\\
Flickr \cite{Flickr}&
      7,575 & 
      239,738 & 
       1,406\\
\hline 
\end{tabular}
\caption{Dataset Statistics}
\label{datastats}
\end{table}

Experiments are implemented in Python using the Pytorch compatible version of DGL \cite{DGL}. In order to remain faithful to the validation splits used in the REDRESS experiments, we split the dataset using 40\% of the edges for train, 40\% for validation, and 20\% for test. For the GCN implementation, we used the hyperparameter settings which had been fine-tuned by Dong et al. During hyperparameter tuning for GraphSAGE, we used grid search to explore learning rates  $\sim$ (0.0001, 0.001, and 0.1), negative sampling of (1, 3 and 5) edges per positive edge, dropout of $\sim$ (0.0, 0.01, 0.03), weight decay of either (0.0, 0.001, or 0.003), warmup epochs of (30, 60, 100, 120) and fairness epochs of (30, 60, 100). Finally, to enable valid comparison with the results attained in REDRESS \cite{Dong_ReDress}, we performed PCA with 200 components to lower the dimensionality of the feature sets before assigning them to each node. 

The best GraphSAGE results for both BlogCatalog and Flickr are achieved using 2 layers of convolutions, a hidden size of $256$, a learning rate of $0.001$, no dropout, $30$ epochs of "warm up" (utility only) training followed by $60$ epochs of utility + fairness training, and a batch size of $32$. In both cases, the best neighborhood size performance was achieved using the smallest neighborhood tests. In the case of BlogCatalog, this was a neighborhood of size 5 (per layer) and in the case of Flickr, this was a neighborhood of size 10 (per layer).

\noindent \textbf{Similarity Notions}: 
We use the cosine similarity metric \cite{cosine_sim} to define both the apriori and prediction matrices $S_G$ and $S_{\hat{Y}}$, respectively. The purpose of selecting this method is to combine the idea of individual characteristics native to individual fairness with a commonly used similarity measure within the recommendation domain. More specifically, cosine similarity is a feature based metric which measures the distance between items in their feature or embedding space. Note that, although this method is flexible to selecting only portions of the feature set to define similarity, in our experiments we use the entire feature set when defining cosine similarity between node pairs. 

Please note that in order to allow for valid comparison between the results listed in the REDRESS paper and the results which we have achieved, we re-run their implementation with a $\gamma = 1$ for both architectures. As such, the results we are listing here do not reflect the results listed in their paper (since they are using $\gamma$ values which have been fine-tuned to balance utility and fairness).

\noindent\textbf{Neighborhood Selection}: We extensively test various neighborhood sizes of GraphSAGE,  motivating the importance of neighborhood selection on fairness promotion. As shown in Table \ref{neighborhood1}, we compare neighborhood sizes of 35, 30, 25, 10, and 5 for subsampling of a 2-hop subgraph in GraphSAGE mini-batch training. 


\noindent \textbf{Evaluation Metric}
In order to evaluate the performance of the two models on each dataset, we use the area under receiver operating curve (AUC) metric. Meanwhile, in order to evaluate the fairness performance, we calculate the NDCG@10 \cite{NDCG} between the apriori similarities, $S_G$ and learned representation similarities, $S_{\hat{Y}}$ averaged over all the nodes (see Appendix \ref{Calculating_NDCG} for more details). 

\section{Results}
We investigate four research questions in our experiments:

\noindent \textbf{RQ1: How does the selection of sub-sampling technique affect fairness promotion?}
As shown in Table \ref{results}, we can see that GraphSAGE not only meets, but exceeds the performance achieved by the GCN method proposed in REDRESS \cite{Dong_ReDress} even before the bias mitigation loop (see Vanilla GraphSAGE vs REDRESS GraphSAGE). For example, when looking at the experiments run on the Flickr data, the GCN method with REDRESS fairness promotion is able to achieve a maximum fairness of 20.38 using Cosine similarity but GraphSAGE, without mitigation, achieves 30.24. This shows that even the use of sub-sampling is beneficial for individual fairness. Similarly on the BlogCatalog dataset, the REDRESS GCN achieves a maximum fairness of 19.64 using Cosine similarity, while Vanilla GraphSAGE achieves 29.95 and REDRESS GraphSAGE achieves 52.02.  Intuitively, we believe that this finding can be attributed to the size of neighborhood which is being used to define individual fairness. In training over an entire graph, we are attempting to fix all the discrepancies in the ranked list between disparate areas of the graph. However, it is possible that the features, or structural elements, affecting fairness have local variations, which are too granular to notice when training over the entire graph. By training over a smaller sub-graph, we are able to use local nuances to fine-tune the embeddings towards individual fairness. See RQ4 for more detailed experiments and explanations. 

\begin{table}[H] 
\centering
\footnotesize
\begin{tabular}{llcccc}
\hline
&& \multicolumn{2}{c}{Cosine}  \\ 
\cmidrule(lr){3-4} 
Data & Model & 
      AUC & 
      Fairness & 
       \\
\hline 
BlogCatalog& 
Vanilla GCN&
      86.75 (-)& 
      16.60 (-)&
      \\ 
& REDRESS GCN & 
      64.15 (-26\%) & 
      19.64 (+18\%) &
      \\
& Vanilla GraphSAGE&
      90.86 (-)& 
      29.95 (-) & 
      \\
& REDRESS GraphSAGE & 
      71.27 (-21\%)& 
      52.02 (+73\%)& 
      \\
\hline 
Flickr& Vanilla GCN&
      86.16 (-)&  
      16.72 (-)& 
      \\ 
&REDRESS GCN & 
      63.31 (-26\%)& 
      20.38 (+21\%) & 
      \\ 
&Vanilla GraphSAGE&
      85.80 (-)& 
      30.24 (-)& 
      \\
&REDRESS GraphSAGE & 
      64.09 (-25\%) & 
      41.51 (+37\%)& 
      \\
      \hline
\end{tabular}
\caption{Results for both BlogCatalog and Flickr. Note: the small values in parenthesis indicating a ($+/-$ \%) are meant to indicate the change between a utility based model (such as Vanilla GCN) and a fairness promoting model (such as REDRESS GCN)}

\label{results}
\end{table}

\noindent \textbf{RQ2: How does the selection of model affect the training time?}
As shown in Table \ref{results}, GraphSAGE is able to achieve improved performance on both utility and fairness metrics. In addition, as shown in Table \ref{epochs}, the mini-batch training protocol of GraphSAGE enables it to train for significantly less epochs than GCN. For example, on Flickr, GCN required 200 epochs of pre-training (utility based) and 100 epochs of integrated fairness training to achieve it maximal results. Meanwhile, on the same dataset, GraphSAGE required only 30 epochs of pre-training and 60 epochs of fairness training. Similarly for BlogCatalog, GCN required 200 epochs of pretraining and 60 epochs of fairness training, while GraphSAGE required only 30 pretraining epochs and 60 fairness training epochs. For the sake of comparison, we selected minimally sized graphs (less then 100K nodes) to allow for a valid comparison between GCN and GraphSAGE. The intractability of training a GCN on a larger graph has been heavily documented among the graph community \cite{sampling_survey1, samplesurvey3}.

\begin{table}[H] 
\centering
\footnotesize
\begin{tabular}{llcccc}
\hline
&& \multicolumn{2}{c}{Cosine}  \\ 
\cmidrule(lr){3-4} \cmidrule(lr){5-6}
Data & Model & 
      Pre-training epochs & 
      Fairness training epochs & 
       \\
\hline 
BlogCatalog& 
GCN&
      200& 
      60&
      \\ 
& 
GraphSAGE&
      30& 
      60&
      \\ 
\hline 
Flickr& 
GCN&
      200& 
      100&
      \\ 
& 
GraphSAGE&
      30& 
      60&
      \\ 
\hline 
\end{tabular}
\caption{Number of Epochs to Achieve Maximal Fairness Performance}
\label{epochs}
\end{table}
\noindent \textbf{RQ3: How does the fairness promotion affect the utility performance of GraphSAGE?}
As shown in Table \ref{results}, we can see that with both GCN and GraphSAGE, the promotion of individual fairness leads to a drop in utility performance. For example, in BlogCatalog, the rounds of cosine similarity based fairness training caused the GCN model's performance to drop by 26\% while GraphSAGE dropped by 21\%. 
Given the formulation of the fairness and utility losses, this balance, between utility and fairness, can be managed by tuning the hyperparameter for $\gamma$ in the loss function (see \ref{fair_training} for more details). We leave the experimentation of hyperparameter tuning for future work. 



\noindent \textbf{RQ4: How does the neighborhood selection affect the ability of GraphSAGE to promote individual fairness?}
\label{neighborhood_discussion}
As shown in the discussion of the previous research questions, there is a noticeable improvement in GraphSAGE over GCN for the promotion of individual fairness. We hypothesize that this differences is rooted in the mini-batch training. GraphSAGE's mini-batch training, uses a sampled sub-graph, while GCN uses the entire graph. We believe that the noticeably smaller neighborhood size used in GraphSAGE updates can allow for better fine-tuning of fairness in the representation learning. This is because the features which affect fairness can potentially differ between disparate areas of the graph. 
In order to test this hypothesis, we perform experiments with 5 neighborhood allocations. As mentioned in \ref{GS},  GraphSAGE selects a neighborhood by randomly sampling \textit{p} neighbors from its surrounding neighborhood. As shown in Table \ref{neighborhood1}, we can see that as the neighborhood size grows, the fairness performance drops. Thus, showing a clear connection between neighborhood size and fairness promotion. 

\begin{table}[h!] 
\centering
\footnotesize
\begin{tabular}{rccccc}
\hline
& \multicolumn{4}{c}{Cosine} & \\ 
\cmidrule(lr){2-6} 
\textbf{REDRESS GraphSAGE} & 
        Nodes in layer 1 & 
        Nodes in layer 2 &
      AUC & 
      Fairness & 
       \\
\hline 
        &5&
        5&
      65.86 & 
      58.04 & 
       \\
&10 & 10 &
      46.24 & 
      56.95 & 
       \\
&25 & 15 & 
      71.27 & 
      52.02&
       \\
&30 & 30 &
      69.32& 
      48.71 & 
       \\
&35 & 35&
     67.17 & 
     46.29 & \\
\hline 
\end{tabular}
\caption{Neighborhood Comparison}
\label{neighborhood1}
\end{table}


\section{Conclusion} 
The advances in the field of GNNs have prompted this architecture to become widely adopted into many graph-structured tasks. In tandem with the proliferation of these models, specifically in the space of recommender systems, has come a growing concern of the exponentiation of biases embedded in their outcomes. Although a growing body of recent work has aimed to implement various notions of fairness within the graph space, the area of individual fairness, has not yet been fully explored. Furthermore, many of the methods proposed in this sub-field remain intractable to the large, real world datasets used in recommender settings. Our work sets out to fill in this gap, by implementing REDRESS, an existing framework for individual fairness, and extending it to GraphSAGE, a scalable GNN architecture. In doing so, we show that the addition of mini-batch training via sub-sampling can significantly improve the promotion of individual fairness. Our findings indicate that this improvement is rooted in the neighborhood selection method, which defines the granularity of local fairness patterns. Due to the prevalence of GNNs in recommender systems, these findings also have ripple effects into the domain of recommendations. Given the shown scalability of this individual fairness notion and the significant ties between GraphSAGE and, its down stream recommmender model, PinSage, we lay the foundation for this individual fairness method to be applied to recommendations. By imposing the need for items which are similar in their initial feature space to remain bounded by this similarity in their embedding space, we believe we can mitigate harmful biases prevalent among recommender systems, such as popularity bias. Our findings show the important connection between individual fairness and neighborhood selection. As such, we believe that this work shows the potential for future implications in field of both GNNs and recommender systems. 

\section*{Acknowledgements}
Funding support for project activities has been partially provided by Canada CIFAR AI Chair, Facebook-MEI award, IVADO grant, and NSERC Discovery Grants program (2021-04378).
\appendix
\section{Calculating NDCG} \label{Calculating_NDCG}
Classically, NDCG is defined as a ratio between Discounted Cumulative Gain (DCG) and Ideal Discounted Cumulative Gain (IDCG) with the Normalized Discounted Cumulative Gain defined as $NDCG = \frac{DCG}{IDCG}$.  \\ 
Following the procedure laid out in REDRESS \cite{Dong_ReDress}, we follow following format for the NDCG calculation. 
Given a node $u$, we generate two ranked lists, $L_G$ and $L_{\hat{Y}}$. Each of these lists is populated with the similarity values from the apriori matrix, $S_G$. The first list, $L_G$, follows an ordering based on $S_G$. Meanwhile the second list, $L_{\hat{Y}}$, follows an ordering based on $S_{\hat{Y}}$. Thus, we define: 
\begin{align}
    IDCG(u) = \sum_{i=0}^{k} \frac{2^{s_{u,v_i}} - 1}{log_2(2 + i)}
\end{align} where ${s}_{u,v_i}$ represents the similarity between nodes $u$ and $v_i$ as defined by the apriori matrix, $S_G$ and the nodes $v_i$ are selected from $L_G$. 
\begin{align}
    DCG(u) = \sum_{i=0}^{k} \frac{2^{{s}_{u, v_i}}-1}{log_2(2 + i)} 
\end{align}\\ where ${s}_{u,v_i}$ represents the similarity between nodes $u$ and $v_i$ as defined by the apriori matrix, $S_G$ and the nodes $v_i$ are selected from $L_{\hat{Y}}$. 
\bibliographystyle{ACM-Reference-Format}
\bibliography{sample-base}


\begin{thebibliography}{46}


\ifx \showCODEN    \undefined \def \showCODEN     #1{\unskip}     \fi
\ifx \showDOI      \undefined \def \showDOI       #1{#1}\fi
\ifx \showISBNx    \undefined \def \showISBNx     #1{\unskip}     \fi
\ifx \showISBNxiii \undefined \def \showISBNxiii  #1{\unskip}     \fi
\ifx \showISSN     \undefined \def \showISSN      #1{\unskip}     \fi
\ifx \showLCCN     \undefined \def \showLCCN      #1{\unskip}     \fi
\ifx \shownote     \undefined \def \shownote      #1{#1}          \fi
\ifx \showarticletitle \undefined \def \showarticletitle #1{#1}   \fi
\ifx \showURL      \undefined \def \showURL       {\relax}        \fi
\providecommand\bibfield[2]{#2}
\providecommand\bibinfo[2]{#2}
\providecommand\natexlab[1]{#1}
\providecommand\showeprint[2][]{arXiv:#2}

\bibitem[Agarap(2018)]%
        {relu}
\bibfield{author}{\bibinfo{person}{Abien~Fred Agarap}.}
  \bibinfo{year}{2018}\natexlab{}.
\newblock \showarticletitle{Deep learning using rectified linear units (relu)}.
\newblock \bibinfo{journal}{\emph{arXiv preprint arXiv:1803.08375}}
  (\bibinfo{year}{2018}).
\newblock


\bibitem[Arduini et~al\mbox{.}(2020)]%
        {arduini_advfairness}
\bibfield{author}{\bibinfo{person}{Mario Arduini}, \bibinfo{person}{Lorenzo
  Noci}, \bibinfo{person}{Federico Pirovano}, \bibinfo{person}{Ce Zhang},
  \bibinfo{person}{Yash~Raj Shrestha}, {and} \bibinfo{person}{Bibek Paudel}.}
  \bibinfo{year}{2020}\natexlab{}.
\newblock \showarticletitle{Adversarial Learning for Debiasing Knowledge Graph
  Embeddings}.
\newblock  (\bibinfo{year}{2020}).
\newblock
\urldef\tempurl%
\url{https://doi.org/10.48550/ARXIV.2006.16309}
\showDOI{\tempurl}


\bibitem[Bauer et~al\mbox{.}(2017)]%
        {bauer_music_2017}
\bibfield{author}{\bibinfo{person}{Christine Bauer}, \bibinfo{person}{Marta
  Kholodylo}, {and} \bibinfo{person}{Christine Strauss}.}
  \bibinfo{year}{2017}\natexlab{}.
\newblock \showarticletitle{Music {Recommender} {Systems} {Challenges} and
  {Opportunities} for {Non}-{Superstar} {Artists}}. In
  \bibinfo{booktitle}{\emph{Bled {eConference}}}.
\newblock


\bibitem[Bose and Hamilton(2019)]%
        {bose19_advFinG}
\bibfield{author}{\bibinfo{person}{Avishek Bose} {and} \bibinfo{person}{William
  Hamilton}.} \bibinfo{year}{2019}\natexlab{}.
\newblock \showarticletitle{Compositional Fairness Constraints for Graph
  Embeddings}. In \bibinfo{booktitle}{\emph{Proceedings of the 36th
  International Conference on Machine Learning}}
  \emph{(\bibinfo{series}{Proceedings of Machine Learning Research},
  Vol.~\bibinfo{volume}{97})}, \bibfield{editor}{\bibinfo{person}{Kamalika
  Chaudhuri} {and} \bibinfo{person}{Ruslan Salakhutdinov}} (Eds.).
  \bibinfo{publisher}{PMLR}, \bibinfo{pages}{715--724}.
\newblock
\urldef\tempurl%
\url{https://proceedings.mlr.press/v97/bose19a.html}
\showURL{%
\tempurl}


\bibitem[Burges et~al\mbox{.}(2006)]%
        {LTR}
\bibfield{author}{\bibinfo{person}{Christopher Burges}, \bibinfo{person}{Robert
  Ragno}, {and} \bibinfo{person}{Quoc Le}.} \bibinfo{year}{2006}\natexlab{}.
\newblock \showarticletitle{Learning to Rank with Nonsmooth Cost Functions}. In
  \bibinfo{booktitle}{\emph{Advances in Neural Information Processing
  Systems}}, \bibfield{editor}{\bibinfo{person}{B.~Sch\"{o}lkopf},
  \bibinfo{person}{J.~Platt}, {and} \bibinfo{person}{T.~Hoffman}} (Eds.),
  Vol.~\bibinfo{volume}{19}. \bibinfo{publisher}{MIT Press}.
\newblock
\urldef\tempurl%
\url{https://proceedings.neurips.cc/paper/2006/file/af44c4c56f385c43f2529f9b1b018f6a-Paper.pdf}
\showURL{%
\tempurl}


\bibitem[Chen et~al\mbox{.}(2020)]%
        {GNN_bias}
\bibfield{author}{\bibinfo{person}{Jiawei Chen}, \bibinfo{person}{Hande Dong},
  \bibinfo{person}{Xiang Wang}, \bibinfo{person}{Fuli Feng},
  \bibinfo{person}{Meng Wang}, {and} \bibinfo{person}{Xiangnan He}.}
  \bibinfo{year}{2020}\natexlab{}.
\newblock \bibinfo{title}{Bias and Debias in Recommender System: A Survey and
  Future Directions}.
\newblock
\newblock
\urldef\tempurl%
\url{https://doi.org/10.48550/ARXIV.2010.03240}
\showDOI{\tempurl}


\bibitem[Chen et~al\mbox{.}(2018)]%
        {FastGCN}
\bibfield{author}{\bibinfo{person}{Jie Chen}, \bibinfo{person}{Tengfei Ma},
  {and} \bibinfo{person}{Cao Xiao}.} \bibinfo{year}{2018}\natexlab{}.
\newblock \bibinfo{title}{FastGCN: Fast Learning with Graph Convolutional
  Networks via Importance Sampling}.
\newblock
\newblock
\urldef\tempurl%
\url{https://doi.org/10.48550/ARXIV.1801.10247}
\showDOI{\tempurl}


\bibitem[Chiang et~al\mbox{.}(2019)]%
        {ClusterGCN}
\bibfield{author}{\bibinfo{person}{Wei-Lin Chiang}, \bibinfo{person}{Xuanqing
  Liu}, \bibinfo{person}{Si Si}, \bibinfo{person}{Yang Li},
  \bibinfo{person}{Samy Bengio}, {and} \bibinfo{person}{Cho-Jui Hsieh}.}
  \bibinfo{year}{2019}\natexlab{}.
\newblock \showarticletitle{Cluster-{GCN}}. In
  \bibinfo{booktitle}{\emph{Proceedings of the 25th {ACM} {SIGKDD}
  International Conference on Knowledge Discovery \& Data Mining}}.
  \bibinfo{publisher}{{ACM}}.
\newblock
\urldef\tempurl%
\url{https://doi.org/10.1145/3292500.3330925}
\showDOI{\tempurl}


\bibitem[Chicaiza and D{\'i}az(2021)]%
        {Chicaiza2021ACS_GNN+RS_SURVEY6}
\bibfield{author}{\bibinfo{person}{Janneth Chicaiza} {and}
  \bibinfo{person}{Priscila~Valdiviezo D{\'i}az}.}
  \bibinfo{year}{2021}\natexlab{}.
\newblock \showarticletitle{A Comprehensive Survey of Knowledge Graph-Based
  Recommender Systems: Technologies, Development, and Contributions}.
\newblock \bibinfo{journal}{\emph{Inf.}}  \bibinfo{volume}{12}
  (\bibinfo{year}{2021}), \bibinfo{pages}{232}.
\newblock


\bibitem[Chierichetti et~al\mbox{.}(2018)]%
        {Chierichetti_fairlets}
\bibfield{author}{\bibinfo{person}{Flavio Chierichetti}, \bibinfo{person}{Ravi
  Kumar}, \bibinfo{person}{Silvio Lattanzi}, {and} \bibinfo{person}{Sergei
  Vassilvitskii}.} \bibinfo{year}{2018}\natexlab{}.
\newblock \showarticletitle{Fair Clustering Through Fairlets}.
\newblock  (\bibinfo{year}{2018}).
\newblock
\urldef\tempurl%
\url{https://doi.org/10.48550/ARXIV.1802.05733}
\showDOI{\tempurl}


\bibitem[Cong et~al\mbox{.}(2021)]%
        {samplesurvey3}
\bibfield{author}{\bibinfo{person}{Weilin Cong}, \bibinfo{person}{Morteza
  Ramezani}, {and} \bibinfo{person}{Mehrdad Mahdavi}.}
  \bibinfo{year}{2021}\natexlab{}.
\newblock \bibinfo{title}{On the Importance of Sampling in Training GCNs:
  Tighter Analysis and Variance Reduction}.
\newblock
\newblock
\urldef\tempurl%
\url{https://doi.org/10.48550/ARXIV.2103.02696}
\showDOI{\tempurl}


\bibitem[Da’u and Salim(2019)]%
        {Dau2019_GNN+RS_SURVEY5}
\bibfield{author}{\bibinfo{person}{Aminu Da’u} {and} \bibinfo{person}{Naomie
  Salim}.} \bibinfo{year}{2019}\natexlab{}.
\newblock \showarticletitle{Recommendation system based on deep learning
  methods: a systematic review and new directions}.
\newblock \bibinfo{journal}{\emph{Artificial Intelligence Review}}
  \bibinfo{volume}{53} (\bibinfo{year}{2019}), \bibinfo{pages}{2709--2748}.
\newblock


\bibitem[Dong et~al\mbox{.}(2021)]%
        {Dong_ReDress}
\bibfield{author}{\bibinfo{person}{Yushun Dong}, \bibinfo{person}{Jian Kang},
  \bibinfo{person}{Hanghang Tong}, {and} \bibinfo{person}{Jundong Li}.}
  \bibinfo{year}{2021}\natexlab{}.
\newblock \showarticletitle{Individual Fairness for Graph Neural Networks: A
  Ranking Based Approach}. In \bibinfo{booktitle}{\emph{Proceedings of the 27th
  ACM SIGKDD Conference on Knowledge Discovery \&; Data Mining}} (Virtual
  Event, Singapore) \emph{(\bibinfo{series}{KDD '21})}.
  \bibinfo{publisher}{Association for Computing Machinery},
  \bibinfo{address}{New York, NY, USA}, \bibinfo{pages}{300–310}.
\newblock
\showISBNx{9781450383325}
\urldef\tempurl%
\url{https://doi.org/10.1145/3447548.3467266}
\showDOI{\tempurl}


\bibitem[Dong et~al\mbox{.}(2022)]%
        {ModelingBias_Dong_2022}
\bibfield{author}{\bibinfo{person}{Yushun Dong}, \bibinfo{person}{Ninghao Liu},
  \bibinfo{person}{Brian Jalaian}, {and} \bibinfo{person}{Jundong Li}.}
  \bibinfo{year}{2022}\natexlab{}.
\newblock \showarticletitle{{EDITS}: Modeling and Mitigating Data Bias for
  Graph Neural Networks}. In \bibinfo{booktitle}{\emph{Proceedings of the {ACM}
  Web Conference 2022}}. \bibinfo{publisher}{{ACM}}.
\newblock
\urldef\tempurl%
\url{https://doi.org/10.1145/3485447.3512173}
\showDOI{\tempurl}


\bibitem[Dwork et~al\mbox{.}(2011)]%
        {Dwork_indfairness}
\bibfield{author}{\bibinfo{person}{Cynthia Dwork}, \bibinfo{person}{Moritz
  Hardt}, \bibinfo{person}{Toniann Pitassi}, \bibinfo{person}{Omer Reingold},
  {and} \bibinfo{person}{Rich Zemel}.} \bibinfo{year}{2011}\natexlab{}.
\newblock \bibinfo{title}{Fairness Through Awareness}.
\newblock
\newblock
\urldef\tempurl%
\url{https://doi.org/10.48550/ARXIV.1104.3913}
\showDOI{\tempurl}


\bibitem[Ekstrand et~al\mbox{.}(2018)]%
        {ekstrand}
\bibfield{author}{\bibinfo{person}{Michael Ekstrand}, \bibinfo{person}{Mucun
  Tian}, \bibinfo{person}{Ion Azpiazu}, \bibinfo{person}{Jennifer Ekstrand},
  \bibinfo{person}{Oghenemaro Anuyah}, \bibinfo{person}{David McNeill}, {and}
  \bibinfo{person}{Maria Pera}.} \bibinfo{year}{2018}\natexlab{}.
\newblock \showarticletitle{All The Cool Kids, How Do They Fit In?: Popularity
  and Demographic Biases in Recommender Evaluation and Effectiveness}.
\newblock


\bibitem[Guo et~al\mbox{.}(2022)]%
        {GNN+RS_SURVEY4}
\bibfield{author}{\bibinfo{person}{Qingyu Guo}, \bibinfo{person}{Fuzhen
  Zhuang}, \bibinfo{person}{Chuan Qin}, \bibinfo{person}{Hengshu Zhu},
  \bibinfo{person}{Xing Xie}, \bibinfo{person}{Hui Xiong}, {and}
  \bibinfo{person}{Qing He}.} \bibinfo{year}{2022}\natexlab{}.
\newblock \showarticletitle{A Survey on Knowledge Graph-Based Recommender
  Systems}.
\newblock \bibinfo{journal}{\emph{IEEE Transactions on Knowledge and Data
  Engineering}} \bibinfo{volume}{34}, \bibinfo{number}{8},
  \bibinfo{pages}{3549--3568}.
\newblock
\urldef\tempurl%
\url{https://doi.org/10.1109/TKDE.2020.3028705}
\showDOI{\tempurl}


\bibitem[Hamilton(2020)]%
        {Will_book}
\bibfield{author}{\bibinfo{person}{William~L. Hamilton}.}
  \bibinfo{year}{2020}\natexlab{}.
\newblock \showarticletitle{Graph Representation Learning}.
\newblock \bibinfo{journal}{\emph{Synthesis Lectures on Artificial Intelligence
  and Machine Learning}} \bibinfo{volume}{14}, \bibinfo{number}{3},
  \bibinfo{pages}{1--159}.
\newblock


\bibitem[Hamilton et~al\mbox{.}(2017)]%
        {Hamilton17_GS}
\bibfield{author}{\bibinfo{person}{William~L. Hamilton}, \bibinfo{person}{Rex
  Ying}, {and} \bibinfo{person}{Jure Leskovec}.}
  \bibinfo{year}{2017}\natexlab{}.
\newblock \bibinfo{title}{Inductive Representation Learning on Large Graphs}.
\newblock
\newblock
\urldef\tempurl%
\url{https://doi.org/10.48550/ARXIV.1706.02216}
\showDOI{\tempurl}


\bibitem[Huang et~al\mbox{.}(2018)]%
        {ASGCN}
\bibfield{author}{\bibinfo{person}{Wenbing Huang}, \bibinfo{person}{Tong
  Zhang}, \bibinfo{person}{Yu Rong}, {and} \bibinfo{person}{Junzhou Huang}.}
  \bibinfo{year}{2018}\natexlab{}.
\newblock \showarticletitle{Adaptive Sampling Towards Fast Graph Representation
  Learning}. In \bibinfo{booktitle}{\emph{Advances in Neural Information
  Processing Systems}}, \bibfield{editor}{\bibinfo{person}{S.~Bengio},
  \bibinfo{person}{H.~Wallach}, \bibinfo{person}{H.~Larochelle},
  \bibinfo{person}{K.~Grauman}, \bibinfo{person}{N.~Cesa-Bianchi}, {and}
  \bibinfo{person}{R.~Garnett}} (Eds.), Vol.~\bibinfo{volume}{31}.
  \bibinfo{publisher}{Curran Associates, Inc.}
\newblock
\urldef\tempurl%
\url{https://proceedings.neurips.cc/paper/2018/file/01eee509ee2f68dc6014898c309e86bf-Paper.pdf}
\showURL{%
\tempurl}


\bibitem[J\"{a}rvelin and Kek\"{a}l\"{a}inen(2002)]%
        {NDCG}
\bibfield{author}{\bibinfo{person}{Kalervo J\"{a}rvelin} {and}
  \bibinfo{person}{Jaana Kek\"{a}l\"{a}inen}.} \bibinfo{year}{2002}\natexlab{}.
\newblock \showarticletitle{Cumulated Gain-Based Evaluation of IR Techniques}.
\newblock \bibinfo{journal}{\emph{ACM Trans. Inf. Syst.}} \bibinfo{volume}{20},
  \bibinfo{number}{4} (\bibinfo{date}{oct} \bibinfo{year}{2002}),
  \bibinfo{pages}{422–446}.
\newblock
\showISSN{1046-8188}
\urldef\tempurl%
\url{https://doi.org/10.1145/582415.582418}
\showDOI{\tempurl}


\bibitem[Kang et~al\mbox{.}(2020)]%
        {Kang_inform}
\bibfield{author}{\bibinfo{person}{Jian Kang}, \bibinfo{person}{Jingrui He},
  \bibinfo{person}{Ross Maciejewski}, {and} \bibinfo{person}{Hanghang Tong}.}
  \bibinfo{year}{2020}\natexlab{}.
\newblock \showarticletitle{InFoRM: Individual Fairness on Graph Mining}. In
  \bibinfo{booktitle}{\emph{Proceedings of the 26th ACM SIGKDD International
  Conference on Knowledge Discovery \&; Data Mining}} (Virtual Event, CA, USA)
  \emph{(\bibinfo{series}{KDD '20})}. \bibinfo{publisher}{Association for
  Computing Machinery}, \bibinfo{address}{New York, NY, USA},
  \bibinfo{pages}{379–389}.
\newblock
\showISBNx{9781450379984}
\urldef\tempurl%
\url{https://doi.org/10.1145/3394486.3403080}
\showDOI{\tempurl}


\bibitem[Khatter et~al\mbox{.}(2021)]%
        {cosine_sim}
\bibfield{author}{\bibinfo{person}{Harsh Khatter}, \bibinfo{person}{Nishtha
  Goel}, \bibinfo{person}{Naina Gupta}, {and} \bibinfo{person}{Muskan Gulati}.}
  \bibinfo{year}{2021}\natexlab{}.
\newblock \showarticletitle{Movie Recommendation System using Cosine Similarity
  with Sentiment Analysis}. In \bibinfo{booktitle}{\emph{2021 Third
  International Conference on Inventive Research in Computing Applications
  (ICIRCA)}}. \bibinfo{pages}{597--603}.
\newblock
\urldef\tempurl%
\url{https://doi.org/10.1109/ICIRCA51532.2021.9544794}
\showDOI{\tempurl}


\bibitem[Kipf and Welling(2016)]%
        {Kipf16_GCN}
\bibfield{author}{\bibinfo{person}{Thomas~N. Kipf} {and} \bibinfo{person}{Max
  Welling}.} \bibinfo{year}{2016}\natexlab{}.
\newblock \bibinfo{title}{Semi-Supervised Classification with Graph
  Convolutional Networks}.
\newblock
\newblock
\urldef\tempurl%
\url{https://doi.org/10.48550/ARXIV.1609.02907}
\showDOI{\tempurl}


\bibitem[Kleinberg et~al\mbox{.}(2016)]%
        {Kleinberg_groupF1}
\bibfield{author}{\bibinfo{person}{Jon Kleinberg}, \bibinfo{person}{Sendhil
  Mullainathan}, {and} \bibinfo{person}{Manish Raghavan}.}
  \bibinfo{year}{2016}\natexlab{}.
\newblock \bibinfo{title}{Inherent Trade-Offs in the Fair Determination of Risk
  Scores}.
\newblock
\newblock
\urldef\tempurl%
\url{https://doi.org/10.48550/ARXIV.1609.05807}
\showDOI{\tempurl}


\bibitem[Kusner et~al\mbox{.}(2017)]%
        {Kusner_counterfactF}
\bibfield{author}{\bibinfo{person}{Matt~J. Kusner}, \bibinfo{person}{Joshua~R.
  Loftus}, \bibinfo{person}{Chris Russell}, {and} \bibinfo{person}{Ricardo
  Silva}.} \bibinfo{year}{2017}\natexlab{}.
\newblock \bibinfo{title}{Counterfactual Fairness}.
\newblock
\newblock
\urldef\tempurl%
\url{https://doi.org/10.48550/ARXIV.1703.06856}
\showDOI{\tempurl}


\bibitem[Lahoti et~al\mbox{.}(2019)]%
        {Lahoti_PFR}
\bibfield{author}{\bibinfo{person}{Preethi Lahoti}, \bibinfo{person}{Krishna~P.
  Gummadi}, {and} \bibinfo{person}{Gerhard Weikum}.}
  \bibinfo{year}{2019}\natexlab{}.
\newblock \showarticletitle{Operationalizing individual fairness with pairwise
  fair representations}.
\newblock \bibinfo{journal}{\emph{Proceedings of the {VLDB} Endowment}}
  \bibinfo{volume}{13}, \bibinfo{number}{4} (\bibinfo{date}{dec}
  \bibinfo{year}{2019}), \bibinfo{pages}{506--518}.
\newblock
\urldef\tempurl%
\url{https://doi.org/10.14778/3372716.3372723}
\showDOI{\tempurl}


\bibitem[Li et~al\mbox{.}(2021a)]%
        {Li_2021_group+RS}
\bibfield{author}{\bibinfo{person}{Yunqi Li}, \bibinfo{person}{Hanxiong Chen},
  \bibinfo{person}{Zuohui Fu}, \bibinfo{person}{Yingqiang Ge}, {and}
  \bibinfo{person}{Yongfeng Zhang}.} \bibinfo{year}{2021}\natexlab{a}.
\newblock \showarticletitle{User-oriented Fairness in Recommendation}. In
  \bibinfo{booktitle}{\emph{Proceedings of the Web Conference 2021}}.
  \bibinfo{publisher}{{ACM}}.
\newblock
\urldef\tempurl%
\url{https://doi.org/10.1145/3442381.3449866}
\showDOI{\tempurl}


\bibitem[Li et~al\mbox{.}(2021b)]%
        {Li_2021_CF+RS}
\bibfield{author}{\bibinfo{person}{Yunqi Li}, \bibinfo{person}{Hanxiong Chen},
  \bibinfo{person}{Shuyuan Xu}, \bibinfo{person}{Yingqiang Ge}, {and}
  \bibinfo{person}{Yongfeng Zhang}.} \bibinfo{year}{2021}\natexlab{b}.
\newblock \showarticletitle{Towards Personalized Fairness based on Causal
  Notion}. In \bibinfo{booktitle}{\emph{Proceedings of the 44th International
  {ACM} {SIGIR} Conference on Research and Development in Information
  Retrieval}}. \bibinfo{publisher}{{ACM}}.
\newblock
\urldef\tempurl%
\url{https://doi.org/10.1145/3404835.3462966}
\showDOI{\tempurl}


\bibitem[Liu et~al\mbox{.}(2021)]%
        {sampling_survey1}
\bibfield{author}{\bibinfo{person}{Xin Liu}, \bibinfo{person}{Mingyu Yan},
  \bibinfo{person}{Lei Deng}, \bibinfo{person}{Guoqi Li},
  \bibinfo{person}{Xiaochun Ye}, {and} \bibinfo{person}{Dongrui Fan}.}
  \bibinfo{year}{2021}\natexlab{}.
\newblock \bibinfo{title}{Sampling methods for efficient training of graph
  convolutional networks: A survey}.
\newblock
\newblock
\urldef\tempurl%
\url{https://doi.org/10.48550/ARXIV.2103.05872}
\showDOI{\tempurl}


\bibitem[Ma et~al\mbox{.}(2022)]%
        {Ma_counterfacFinG}
\bibfield{author}{\bibinfo{person}{Jing Ma}, \bibinfo{person}{Ruocheng Guo},
  \bibinfo{person}{Mengting Wan}, \bibinfo{person}{Longqi Yang},
  \bibinfo{person}{Aidong Zhang}, {and} \bibinfo{person}{Jundong Li}.}
  \bibinfo{year}{2022}\natexlab{}.
\newblock \showarticletitle{Learning Fair Node Representations with Graph
  Counterfactual Fairness}. In \bibinfo{booktitle}{\emph{Proceedings of the
  Fifteenth {ACM} International Conference on Web Search and Data Mining}}.
  \bibinfo{publisher}{{ACM}}.
\newblock
\urldef\tempurl%
\url{https://doi.org/10.1145/3488560.3498391}
\showDOI{\tempurl}


\bibitem[Mehrotra et~al\mbox{.}(2018)]%
        {marketplace}
\bibfield{author}{\bibinfo{person}{Rishabh Mehrotra}, \bibinfo{person}{James
  McInerney}, \bibinfo{person}{Hugues Bouchard}, \bibinfo{person}{Mounia
  Lalmas}, {and} \bibinfo{person}{Fernando Diaz}.}
  \bibinfo{year}{2018}\natexlab{}.
\newblock \showarticletitle{Towards a Fair Marketplace: Counterfactual
  Evaluation of the Trade-off between Relevance, Fairness \&; Satisfaction in
  Recommendation Systems}. In \bibinfo{booktitle}{\emph{Proceedings of the 27th
  ACM International Conference on Information and Knowledge Management}}
  (Torino, Italy) \emph{(\bibinfo{series}{CIKM '18})}.
  \bibinfo{publisher}{Association for Computing Machinery},
  \bibinfo{address}{New York, NY, USA}, \bibinfo{pages}{2243–2251}.
\newblock
\showISBNx{9781450360142}
\urldef\tempurl%
\url{https://doi.org/10.1145/3269206.3272027}
\showDOI{\tempurl}


\bibitem[Quadrana et~al\mbox{.}(2018)]%
        {GNN+RS_SURVEY3}
\bibfield{author}{\bibinfo{person}{Massimo Quadrana}, \bibinfo{person}{Paolo
  Cremonesi}, {and} \bibinfo{person}{Dietmar Jannach}.}
  \bibinfo{year}{2018}\natexlab{}.
\newblock \showarticletitle{Sequence-Aware Recommender Systems}.
\newblock \bibinfo{journal}{\emph{ACM Comput. Surv.}} \bibinfo{volume}{51},
  \bibinfo{number}{4}, Article \bibinfo{articleno}{66},
  \bibinfo{numpages}{36}~pages.
\newblock
\showISSN{0360-0300}
\urldef\tempurl%
\url{https://doi.org/10.1145/3190616}
\showDOI{\tempurl}


\bibitem[Tang and Liu(2009a)]%
        {Blog_Catalog}
\bibfield{author}{\bibinfo{person}{Lei Tang} {and} \bibinfo{person}{Huan Liu}.}
  \bibinfo{year}{2009}\natexlab{a}.
\newblock \showarticletitle{Relational Learning via Latent Social Dimensions}.
  In \bibinfo{booktitle}{\emph{Proceedings of the 15th ACM SIGKDD International
  Conference on Knowledge Discovery and Data Mining}} (Paris, France)
  \emph{(\bibinfo{series}{KDD '09})}. \bibinfo{publisher}{Association for
  Computing Machinery}, \bibinfo{address}{New York, NY, USA},
  \bibinfo{pages}{817–826}.
\newblock
\showISBNx{9781605584959}
\urldef\tempurl%
\url{https://doi.org/10.1145/1557019.1557109}
\showDOI{\tempurl}


\bibitem[Tang and Liu(2009b)]%
        {Flickr}
\bibfield{author}{\bibinfo{person}{Lei Tang} {and} \bibinfo{person}{Huan Liu}.}
  \bibinfo{year}{2009}\natexlab{b}.
\newblock \showarticletitle{Relational Learning via Latent Social Dimensions}.
  In \bibinfo{booktitle}{\emph{Proceedings of the 15th ACM SIGKDD International
  Conference on Knowledge Discovery and Data Mining}} (Paris, France)
  \emph{(\bibinfo{series}{KDD '09})}. \bibinfo{publisher}{Association for
  Computing Machinery}, \bibinfo{address}{New York, NY, USA},
  \bibinfo{pages}{817–826}.
\newblock
\showISBNx{9781605584959}
\urldef\tempurl%
\url{https://doi.org/10.1145/1557019.1557109}
\showDOI{\tempurl}


\bibitem[Wang et~al\mbox{.}(2019)]%
        {DGL}
\bibfield{author}{\bibinfo{person}{Minjie Wang}, \bibinfo{person}{Da Zheng},
  \bibinfo{person}{Zihao Ye}, \bibinfo{person}{Quan Gan},
  \bibinfo{person}{Mufei Li}, \bibinfo{person}{Xiang Song},
  \bibinfo{person}{Jinjing Zhou}, \bibinfo{person}{Chao Ma},
  \bibinfo{person}{Lingfan Yu}, \bibinfo{person}{Yu Gai},
  \bibinfo{person}{Tianjun Xiao}, \bibinfo{person}{Tong He},
  \bibinfo{person}{George Karypis}, \bibinfo{person}{Jinyang Li}, {and}
  \bibinfo{person}{Zheng Zhang}.} \bibinfo{year}{2019}\natexlab{}.
\newblock \bibinfo{title}{Deep Graph Library: A Graph-Centric,
  Highly-Performant Package for Graph Neural Networks}.
\newblock
\newblock
\urldef\tempurl%
\url{https://doi.org/10.48550/ARXIV.1909.01315}
\showDOI{\tempurl}


\bibitem[Wang et~al\mbox{.}(2021)]%
        {sampling_survey2}
\bibfield{author}{\bibinfo{person}{Zhaokang Wang}, \bibinfo{person}{Yunpan
  Wang}, \bibinfo{person}{Chunfeng Yuan}, \bibinfo{person}{Rong Gu}, {and}
  \bibinfo{person}{Yihua Huang}.} \bibinfo{year}{2021}\natexlab{}.
\newblock \showarticletitle{Empirical analysis of performance bottlenecks in
  graph neural network training and inference with GPUs}.
\newblock \bibinfo{journal}{\emph{Neurocomputing}}  \bibinfo{volume}{446}
  (\bibinfo{year}{2021}), \bibinfo{pages}{165--191}.
\newblock
\showISSN{0925-2312}
\urldef\tempurl%
\url{https://doi.org/10.1016/j.neucom.2021.03.015}
\showDOI{\tempurl}


\bibitem[Wu et~al\mbox{.}(2021a)]%
        {Wu2021MultiFRAM}
\bibfield{author}{\bibinfo{person}{Haolun Wu}, \bibinfo{person}{Chen Ma},
  \bibinfo{person}{Bhaskar Mitra}, \bibinfo{person}{Fernando Diaz}, {and}
  \bibinfo{person}{Xue Liu}.} \bibinfo{year}{2021}\natexlab{a}.
\newblock \showarticletitle{Multi-FR: A Multi-Objective Optimization Method for
  Achieving Two-sided Fairness in E-commerce Recommendation}.
\newblock \bibinfo{journal}{\emph{ArXiv}}  \bibinfo{volume}{abs/2105.02951}
  (\bibinfo{year}{2021}).
\newblock


\bibitem[Wu et~al\mbox{.}(2020)]%
        {GNN_in_RecSys}
\bibfield{author}{\bibinfo{person}{Shiwen Wu}, \bibinfo{person}{Fei Sun},
  \bibinfo{person}{Wentao Zhang}, \bibinfo{person}{Xu Xie}, {and}
  \bibinfo{person}{Bin Cui}.} \bibinfo{year}{2020}\natexlab{}.
\newblock \bibinfo{title}{Graph Neural Networks in Recommender Systems: A
  Survey}.
\newblock
\newblock
\urldef\tempurl%
\url{https://doi.org/10.48550/ARXIV.2011.02260}
\showDOI{\tempurl}


\bibitem[Wu et~al\mbox{.}(2021b)]%
        {GNN_survey}
\bibfield{author}{\bibinfo{person}{Zonghan Wu}, \bibinfo{person}{Shirui Pan},
  \bibinfo{person}{Fengwen Chen}, \bibinfo{person}{Guodong Long},
  \bibinfo{person}{Chengqi Zhang}, {and} \bibinfo{person}{Philip~S. Yu}.}
  \bibinfo{year}{2021}\natexlab{b}.
\newblock \showarticletitle{A Comprehensive Survey on Graph Neural Networks}.
\newblock \bibinfo{journal}{\emph{IEEE Transactions on Neural Networks and
  Learning Systems}} \bibinfo{volume}{32}, \bibinfo{number}{1}
  (\bibinfo{date}{Jan} \bibinfo{year}{2021}), \bibinfo{pages}{4–24}.
\newblock
\showISSN{2162-2388}
\urldef\tempurl%
\url{https://doi.org/10.1109/tnnls.2020.2978386}
\showDOI{\tempurl}


\bibitem[Ying et~al\mbox{.}(2018)]%
        {pinSAGE}
\bibfield{author}{\bibinfo{person}{Rex Ying}, \bibinfo{person}{Ruining He},
  \bibinfo{person}{Kaifeng Chen}, \bibinfo{person}{Pong Eksombatchai},
  \bibinfo{person}{William~L. Hamilton}, {and} \bibinfo{person}{Jure
  Leskovec}.} \bibinfo{year}{2018}\natexlab{}.
\newblock \showarticletitle{Graph Convolutional Neural Networks for Web-Scale
  Recommender Systems}. In \bibinfo{booktitle}{\emph{Proceedings of the 24th
  {ACM} {SIGKDD} International Conference on Knowledge Discovery \& Data
  Mining}}. \bibinfo{publisher}{{ACM}}.
\newblock
\urldef\tempurl%
\url{https://doi.org/10.1145/3219819.3219890}
\showDOI{\tempurl}


\bibitem[Zeng et~al\mbox{.}(2019)]%
        {GraphSAINT}
\bibfield{author}{\bibinfo{person}{Hanqing Zeng}, \bibinfo{person}{Hongkuan
  Zhou}, \bibinfo{person}{Ajitesh Srivastava}, \bibinfo{person}{Rajgopal
  Kannan}, {and} \bibinfo{person}{Viktor Prasanna}.}
  \bibinfo{year}{2019}\natexlab{}.
\newblock \bibinfo{title}{GraphSAINT: Graph Sampling Based Inductive Learning
  Method}.
\newblock
\newblock
\urldef\tempurl%
\url{https://doi.org/10.48550/ARXIV.1907.04931}
\showDOI{\tempurl}


\bibitem[Zhang et~al\mbox{.}(2020)]%
        {GNN+RS_SURVEY2}
\bibfield{author}{\bibinfo{person}{Shuai Zhang}, \bibinfo{person}{Lina Yao},
  \bibinfo{person}{Aixin Sun}, {and} \bibinfo{person}{Yi Tay}.}
  \bibinfo{year}{2020}\natexlab{}.
\newblock \showarticletitle{Deep Learning Based Recommender System}.
\newblock \bibinfo{journal}{\emph{Comput. Surveys}} \bibinfo{volume}{52},
  \bibinfo{number}{1} (\bibinfo{date}{jan} \bibinfo{year}{2020}),
  \bibinfo{pages}{1--38}.
\newblock
\urldef\tempurl%
\url{https://doi.org/10.1145/3285029}
\showDOI{\tempurl}


\bibitem[Zhang et~al\mbox{.}(2021)]%
        {Zhang_2021_C+RS2}
\bibfield{author}{\bibinfo{person}{Yang Zhang}, \bibinfo{person}{Fuli Feng},
  \bibinfo{person}{Xiangnan He}, \bibinfo{person}{Tianxin Wei},
  \bibinfo{person}{Chonggang Song}, \bibinfo{person}{Guohui Ling}, {and}
  \bibinfo{person}{Yongdong Zhang}.} \bibinfo{year}{2021}\natexlab{}.
\newblock \showarticletitle{Causal Intervention for Leveraging Popularity Bias
  in Recommendation}. In \bibinfo{booktitle}{\emph{Proceedings of the 44th
  International {ACM} {SIGIR} Conference on Research and Development in
  Information Retrieval}}. \bibinfo{publisher}{{ACM}}.
\newblock
\urldef\tempurl%
\url{https://doi.org/10.1145/3404835.3462875}
\showDOI{\tempurl}


\bibitem[Zhu et~al\mbox{.}(2021)]%
        {popularity_Bias}
\bibfield{author}{\bibinfo{person}{Ziwei Zhu}, \bibinfo{person}{Yun He},
  \bibinfo{person}{Xing Zhao}, \bibinfo{person}{Yin Zhang},
  \bibinfo{person}{Jianling Wang}, {and} \bibinfo{person}{James Caverlee}.}
  \bibinfo{year}{2021}\natexlab{}.
\newblock \showarticletitle{Popularity-Opportunity Bias in Collaborative
  Filtering}. In \bibinfo{booktitle}{\emph{Proceedings of the 14th ACM
  International Conference on Web Search and Data Mining}} (Virtual Event,
  Israel) \emph{(\bibinfo{series}{WSDM '21})}. \bibinfo{publisher}{Association
  for Computing Machinery}, \bibinfo{address}{New York, NY, USA},
  \bibinfo{pages}{85–93}.
\newblock
\showISBNx{9781450382977}
\urldef\tempurl%
\url{https://doi.org/10.1145/3437963.3441820}
\showDOI{\tempurl}


\bibitem[Zou et~al\mbox{.}(2019)]%
        {LADIES}
\bibfield{author}{\bibinfo{person}{Difan Zou}, \bibinfo{person}{Ziniu Hu},
  \bibinfo{person}{Yewen Wang}, \bibinfo{person}{Song Jiang},
  \bibinfo{person}{Yizhou Sun}, {and} \bibinfo{person}{Quanquan Gu}.}
  \bibinfo{year}{2019}\natexlab{}.
\newblock \showarticletitle{Layer-Dependent Importance Sampling for Training
  Deep and Large Graph Convolutional Networks}.
\newblock


\end{thebibliography}










\end{document}